\documentclass[letterpaper, 10 pt, conference]{ieeeconf}  

\IEEEoverridecommandlockouts                              

\usepackage[utf8]{inputenc}
\usepackage{amsmath,caption,subcaption,graphicx,color}
\usepackage{multirow}

\newenvironment{flushitemize}{%
	\begin{list}{$\bullet$}
		{\setlength{\leftmargin}{15pt}}%
		\setlength{\labelwidth}{20pt}
		\setlength{\itemindent}{0pt}
		\setlength{\labelsep}{0.5em}
		\setlength{\itemsep}{1pt}
		\setlength{\parskip}{0pt}
		\setlength{\parsep}{0pt}}
	{\end{list}}
 
\title{\LARGE \bf
Anticipatory Human-Robot Collaboration via Multi-Objective Trajectory Optimization
}

\author{Abhinav Jain$^{1}$, Daphne Chen$^{1}$, Dhruva Bansal$^{1}$, Sam Scheele$^{1}$, Mayank Kishore$^{1}$, Hritik Sapra$^{1}$, David Kent$^{1}$, \\ Harish Ravichandar$^{1}$, Sonia Chernova$^{1}$
\thanks{*This work was supported in part by the Army Research Lab under Grant W911NF-
17-2-0181 (DCIST CRA).}
\thanks{$^{1}$The authors are with the Institute of Robotics and Intelligent Machines, Georgia Institute of Technology, Atlanta, GA, USA {\tt\small \{jain, daphne.chen, dbansal36, scheele, mkishore3, hritiksapra, dekent, harish.ravichandar, chernova\}@gatech.edu}}
}

\begin{document}

\maketitle

\thispagestyle{empty}
\pagestyle{empty}

\begin{abstract}

  We address the problem of adapting robot trajectories to improve safety, comfort, and efficiency in human-robot collaborative tasks. To this end, we propose \emph{CoMOTO}, a trajectory optimization framework that utilizes stochastic motion prediction to anticipate the human's motion and adapt the robot's joint trajectory accordingly. We design a multi-objective cost function that simultaneously optimizes for \textit{i)} separation distance, \textit{ii)} visibility of the end-effector, \textit{iii)} legibility, \textit{iv)} efficiency, and \textit{v)} smoothness. We evaluate CoMOTO against three existing methods for robot trajectory generation when in close proximity to humans. Our experimental results indicate that our approach consistently outperforms existing methods over a combined set of safety, comfort, and efficiency metrics.

\end{abstract}

\section{Introduction}
\label{sec:introduction}

With the advances in robotics automation, robots are more frequently working in close proximity to humans. For instance, in manufacturing, humans and robots can work together to assemble components, and in household and assistive robotics, robots can provide physical assistance to humans. Thus, there is a growing need for robots to effectively and safely interact with humans in close proximity.

A key challenge in robot-human close-proximity interaction is the generation of robot trajectories that are safe, i.e. they do not physically harm the human, and are comfortable, i.e. the human is able to interpret and anticipate the robot's behavior. Collaborative robotic systems can create safe trajectories with frequent monitoring and replanning~\cite{lasota2014toward, dumonteil2015reactive}, but at the cost of efficiency. Anticipatory methods that use predictions of human motion can instead be used to generate safer trajectories using learned models of human motion.

Several different factors define safety and comfort of a robot's trajectory. While a trajectory may be safe for a nearby human, it might not be comfortable. The visibility of the robot's end effector in the peripheral vision of the human can increase comfort. Inference of the robot's intent through partial observation of its trajectory can also greatly increase comfort. Finally, sudden and unexpected robot behavior can be a major source for discomfort. Effective robot trajectories in human-robot collaboration must take into account all of these factors.  Further, user experience factors are not necessarily complimentary to trajectory efficiency, as previously shown for robot-human handover tasks \cite{huang2015adaptive}.  Thus, collaborative robot trajectory generation must effectively \textit{balance} efficiency, safety, and comfort factors.

In this paper, we address the problem of adapting robot trajectories for human-robot collaborative environments with an overall goal of improving human safety and comfort, as well as increasing task efficiency. We define the default trajectory executed by the robot for a particular task as the nominal trajectory. Given some observation of human motion directly preceding execution of the nominal trajectory, we use a prediction of the human's motion to adapt the nominal trajectory for improved safety and comfort.

\begin{figure}[t]
  \centering
  \includegraphics[width=0.8\columnwidth]{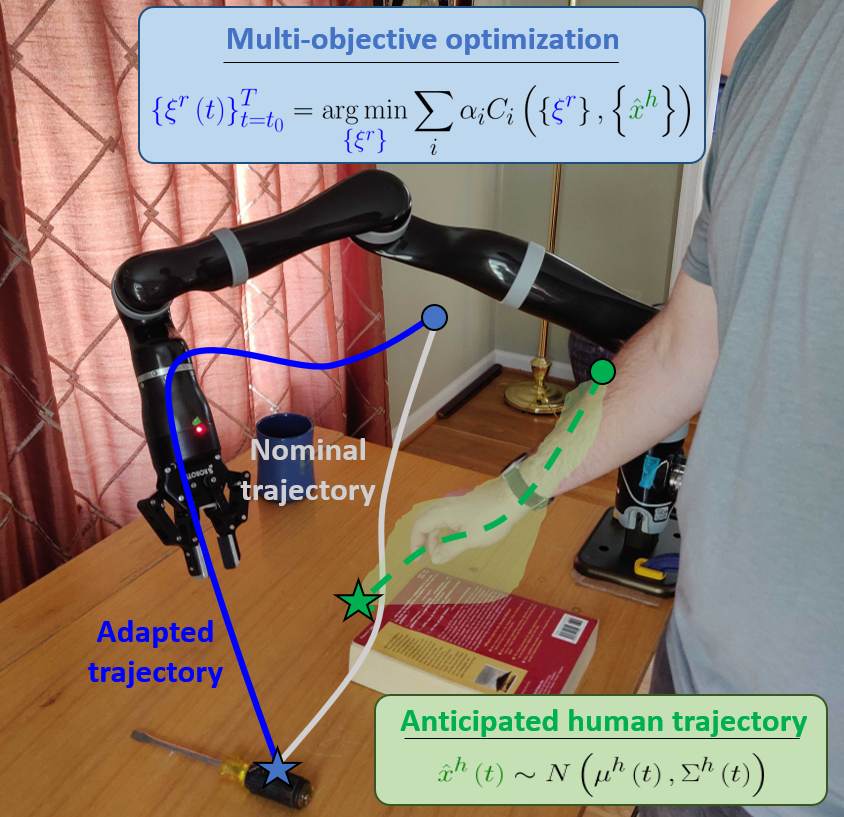}
  \caption{Effective human-robot collaboration requires both anticipation and simultaneous optimization of a variety of costs associated with safety, efficiency, and comfort.}
  \label{fig:overview}
  \vspace{-0.2cm}
\end{figure}

Figure \ref{fig:overview} shows an overview of our approach. We combine multiple objective functions to satisfy several factors of human comfort in addition to safety. We use time-sampled stochastic predictions of human motion $\hat{x}^h\left(t\right)$ to generate objective functions $C_i\left( \xi^r, \hat{x}^h \right)$, using the uncertainty of the prediction to generate appropriately conservative trajectories.  We call our approach Collaborative Multi-Objective Trajectory Optimization -- \emph{CoMOTO}.

In order to evaluate CoMOTO, we define several metrics that incorporate key factors in safety and comfort of humans in collaborative environments. We perform experiments in three collaborative picking test cases, and compare the results against established baselines.
Our results show that CoMOTO performs consistently well for all of our metrics across different close-proximity collaborative picking scenarios, while the baselines are able to perform well in only a particular metric or for only a specific test scenario.

\section{Related Works}
\label{sec:relatedworks}

Human-robot collaborative manipulation is a well-studied topic. Several works focus on reactive systems for safety in collaborative environments. Lasota \textit{et al.} propose a reactive speed control system for collaborative robots~\cite{lasota2014toward}. Their system monitors human pose to measure the separation distance between the human and the robot, scales the robot's execution speed by the separation distance, and stops execution completely below a specified separation threshold. Dumonteil \textit{et al.} propose a similar reactive approach for collaborative robots in industrial applications using a state machine~\cite{dumonteil2015reactive}, by reactively replanning trajectories to avoid potential collisions. Reactive systems based on unadapted trajectories are task inefficient due to repeated replanning, and so we instead focus on using human motion prediction to generate initially safer trajectories.

Several prior works have utilized human motion prediction in order to proactively adapt robot trajectories. Mainprice and Berenson propose a prediction based planning framework that generates swept volumes for collision detection based on human motion prediction using a Gaussian mixture model (GMM), and interleaves planning and execution to update the prediction~\cite{mainprice2013human}. However, since their system selects an updated predicted human trajectory at each iteration, their framework requires constant replanning.
Fishman \textit{et al.} address the problem of coordinated human-robot collaboration, specifically in a handover task~\cite{fishman2019trajectory}, using a joint optimal control model to simultaneously plan the robot's behavior and predict the humans' behavior by inferring human goals.  Stouraitis \textit{et al.} develop a method that involves estimating a human partner's policy to optimize trajectories for dyadic collaborative manipulation, where a human and robot work together to manipulate a single large object \cite{stouraitis2018dyadic}.
Huang and Mutlu present an anticipatory control method based on inference from human gaze~\cite{huang2016anticipatory}, highlighting the task efficiency benefits of using predictive and anticipatory planning methods.
Maeda \textit{et al.} use early human action recognition to initiate a corresponding robot response~\cite{maeda2016anticipative}. Maeda \textit{et al.} also present a Probabilistic Movement Primitive framework for learning a mixture model of human-robot interaction primitives, used to identify human tasks as well as to coordinate robot movement with the observed human movement. These works either focus on improving task efficiency through predicting human intent, or use human motion prediction to create safe trajectories.  Our work builds on these ideas by using human motion prediction to improve human comfort factors as well.


Our objective of combining several factors of safety and comfort for trajectory optimization is similar to Mainprice \textit{et al.}'s work~\cite{mainprice2011planning}, which considers distance for safety, and visibility and reachability for comfort in robot handover tasks. However, their work assumes the human will remain still during the robot's trajectory execution, and it does not consider the pose of the human's arm. We extend collaborative multi-objective trajectory optimization to account for a moving human, with an articulated human model.


Human comfort factors beyond collision avoidance are important considerations in human-robot collaborative environments~\cite{lasota2017survey}. Dragan \textit{et al.} propose legibility and predictability of robot motion to a human observer~\cite{dragan2013legibility, dragan2013generating}. A legible motion is one from which an observer can quickly and confidently infer the motion's goal after only partial observation, and predictable motion is the most expected motion to reach a goal. 
Stulp \textit{et al.} present legibility as a task-specific behavior that can be learned rather than a general characteristic of a trajectory~\cite{stulp2015facilitating}.  Medina \textit{et al.} emphasize the importance of smoothness for robot-human handover trajectories \cite{medina2016human}.  In order to account for multiple factors that affect human comfort, our work considers legibility, predictability (through efficient execution), and smoothness.

\section{Collaborative Trajectory Optimization}
\label{sec:description}

Our framework, CoMOTO, uses stochastic human motion prediction to calculate an objective function, composed of a set of costs relevant to close-proximity interaction, which is minimized using a trajectory optimization framework.  Specifically, the trajectory generation pipeline consists of a brief 1 second observation period of the human's motion, which is used to predict the remaining trajectory of the human (see Section \ref{sec:prediction} for details).  The predicted trajectory is then used as input to calculate a set of costs, including separation distance, visibility, legibility, deviation from a nominal trajectory, and smoothness, that account for the stochastic nature of the prediction.  The costs themselves are detailed in Section \ref{sec:costs}.  We formulate an objective function using a weighted combination of the costs, which is then minimized to generate a robot trajectory using TrajOpt~\cite{schulman2014motion}, although other trajectory optimization works, such as \cite{zucker2013chomp, kalakrishnan2011stomp}, and cost based planning algorithms, such as \cite{jaillet2008transition}, can be used instead. The generated trajectory is then executed concurrently with the remainder of the human's motion.

While sensor-based reactive stops offer absolute collision prevention, they reduce efficiency with frequent interruptions requiring replanning. Instead, we directly address safety during planning. CoMOTO leverages motion prediction to anticipate human motion and  generate trajectories that are inherently safer, thereby reducing the need for replanning and increasing task efficiency.

\section{Trajectory Adaptation Costs}
\label{sec:costs}

Our objective function is split into several costs that cover different elements of safety and comfort in a collaborate environment. Each cost is a function of time parameterized robot joint trajectory $\xi^r \left(t\right)$ and predicted human motion $\hat{x}^h\left(t\right) \sim N\left( \mu^{h} \left(t\right),\Sigma^h\left(t\right) \right)$ for $t \in \left[ 0, T \right]$.

\subsection{Distance Cost}

Distance between the human and the robot is the most critical factor in safe collaborative manipulation. Thus we formulate a cost that penalizes lower separation distances between the human and the robot. The cost is further scaled by the covariance of the prediction, with higher covariance resulting in a higher cost, resulting in more conservative trajectories when the predicted motion has higher uncertainty. The cost is formulated as follows: 
\begin{equation}
\begin{split}
  C_{dist}&\left( \xi^r, \hat{x}^h \right) = \sum_{t = 0}^{T} \sum_{i} \sum_{j}  \frac{1}{d_{i, j}\left(t\right)^T \left(\Sigma_{i}^{h}\left(t\right)\right)^{-1} d_{i, j}\left(t\right)} \\ 
  &d_{i, j}\left(t\right) = \mu_{i}^{h}\left(t\right) - p_{j}^{r}\left(t\right)
\end{split}
\end{equation}
where $\mu_{i}^{h}\left(t\right)$ and $ \Sigma_{i}^{h}\left(t\right)$ are the mean and covariance of the predicted 3D position of the $i^{\text{th}}$ human joint at time $t$, and $p_{j}^{r}\left(t\right)$ is the 3D position of the $j^{\text{th}}$ robot joint at time $t$. 

\subsection{Visibility Cost}

During trajectory execution, visibility of the robot's end effector is an essential factor for human comfort \cite{sisbot2007spatial}. If the robot is out of the field of view of the human, the human may be distracted and try to locate it, thus decreasing both human comfort and task efficiency. This is a basic human instinct for safety against unpredictable moving objects. The visibility cost penalizes the end effector for being farther from the human's gaze.

We define the visibility cost as the angle between the predicted human gaze and the line between the position of the robot end effector and the human's head. We define the predicted human gaze as the line from the predicted position of the human head to the position of the object with which the human is interacting. The cost is scaled inversely to the variance of the prediction of the human head pose. 
\begin{equation}
  C_{vis}\left( \xi^r, \hat{x}^h \right) = \sum_{t = 0}^{T} \frac{ \angle \left(O, \mu_{head}^{h}\left(t\right), p_{eef}^{r}\left(t\right)\right)} { \sigma_{head}^{h}\left(t\right)}
\end{equation}
where $O$ is the 3D position of the object with which the human is interacting, $\mu_{head}^{h}\left(t\right)$ and $\sigma_{head}^{h}\left(t\right)$ are the mean and variance of the predicted 3D position of the human head, and $p_{eef}^{r}\left(t\right)$ is the 3D end-effector position at time $t$.

\subsection{Legibility Cost}

The robot's motion must be legible, that is, it must convey its intent through its trajectory. Dragan \textit{et al.} define a legible robot trajectory as one from which the user can quickly and confidently infer the task goal after only partial trajectory execution \cite{dragan2013legibility}. We choose to implement a legibility cost in order to improve the human's ability to understand the robot's intent. We replicate the Legibility cost from \cite{dragan2013generating}.
\begin{equation}
  C_{legibility}\left( \xi^r \right) = \frac{\sum_{t} P\left(G \vert \xi_{S \to Q_t}^r \right) f\left(t\right)}{\sum_{t} f\left(t\right)}
\end{equation}
\begin{equation}
  P\left(G | \xi_{S \to Q}^r\right) = \frac{\exp\left(-C\left(\xi_{S \to Q}^r\right) - C\left(\xi^{*r}_{Q \to G}\right)\right)}{\exp\left(-C\left(\xi^{*r}_{S \to G}\right)\right)}
\label{eqn:legibility_prob}
\end{equation}
$\xi_{A \to B}^r$ is the trajectory from configuration $A$ to configuration $B$. $S$ denotes the robot's start configuration, $G$ denotes its goal configuration, and $Q_t$ denotes its configuration at time $t$. $f\left(t\right)$ is a weighing function that increases cost of legibility towards the beginning of the trajectory. The optimal trajectory $\xi^{*r}$ is a linear trajectory in the Cartesian space. $C\left(\xi\right)$ is the length of the trajectory in Cartesian space. Dragan \textit{et al.} include a regularizer term $\lambda C\left(\xi\right)$ in order to prevent excessively long trajectories, which we exclude from our cost as that requirement is met by the Nominal Trajectory Cost.

\subsection{Nominal Trajectory Cost}

The nominal trajectory is the default trajectory executed by the robot without any adaptation. This trajectory is calculated using a collision cost and a joint velocity cost in TrajOpt. The nominal trajectory can be viewed as one that optimizes smoothness, collision avoidance (with objects), and efficiency in the \textit{absence} of a human. While the costs defined thus far focus solely on the human, our nominal trajectory cost brings balance to the overall cost function, and acts as a regularizer to preserve efficiency.

The nominal trajectory cost penalizes deviation from the nominal trajectory. The cost is calculated as a sum of Cartesian distances of the end effector between the nominal trajectory and the adapted trajectory at each timestep:
\begin{equation}
    C_{nominal} \left( \xi^r \right) = \sum_{t=0}^{T}\left\Vert p_{nom}^r\left(t\right)  - p_{eef}^r\left(t\right) \right\Vert
\end{equation}
where $p_{nom}^r\left(t\right)$ is the position of the end effector at time $t$ in the nominal trajectory.

\subsection{Smoothness Cost}

Smooth robot motion is a necessary component for a comfortable collaborative environment. Several dynamical quantities can be minimized across the trajectory to generate smooth motion. Prior trajectory optimization frameworks such as \cite{zucker2013chomp} use sum of squared velocities of the robot as a smoothing cost. However, in order to better decrease jerkiness of adapted trajectories as well as to even out speed across the execution of the trajectory, we use the sum of squared acceleration of the robot as follows:
\begin{equation}
    C_{smooth}\left( \xi^r \right) = \sum_{t=0}^{T-2} \left( \left\Vert \frac{d^2}{dt^2}\xi\left(t\right) \right\Vert^2 \right)
\end{equation}

\subsection{Multi-Objective Optimization}

The final objective function is the sum of all the above costs. The overall optimization problem is given by 
\begin{align}
    \{\xi^{r^*}\}_{t=0}^{T} =& \underset{x}{\arg \min} \sum_{i \in \mathcal{C}} \alpha_i C_{i} \left( \xi^r, \hat{x}^h \right) \\
     & \quad \mathrm{s.t.} \quad \xi^{r^*}(T) = \xi^{r}_{g} 
\end{align}
where $\{\xi^{r^*}\}_{t=0}^{T}$ denotes the optimized robot joint trajectory, $\mathcal{C}=\{dist, vis, legibility, nominal, smooth\}$, $\alpha_i$ represents the pre-specified weights associated with the $i^{\text{th}}$ cost, and $\xi^{r}_{g}$ denotes the desired goal location.

We note that the costs used in the objective function do not necessarily incentivize the same behavior. For instance, minimizing the distance cost will push the robot trajectory away from the human. On the other hand, the visibility cost will work to pull the trajectory closer to the human. Nevertheless, each cost function considers an important aspect of the interaction. CoMOTO thus attempts to find optimized trajectories that effectively trade-off different costs according to the specified weighting factors.

\section{Experiments and Results}
\label{sec:results}

\begin{figure}[t]
    \centering
    \includegraphics[width=0.75\columnwidth]{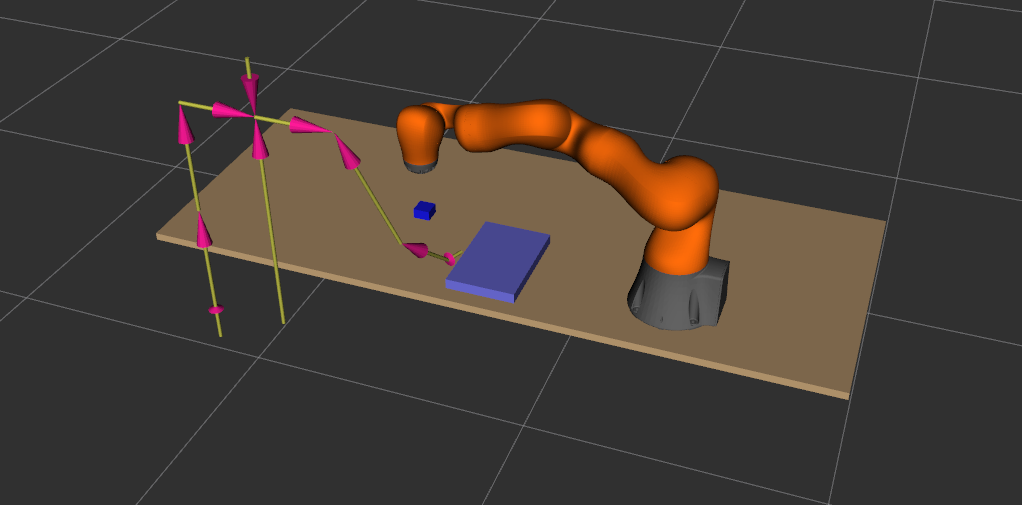}
    \caption{Simulated test environment showing the KUKA robot and the keypoints of the human skeleton.}
    \label{fig:environment}
    \vspace{-0.2cm}
\end{figure}

We perform a series of experiments on three different test cases involving a human and a robot working in a collaborative environment in order to evaluate the performance of our approach. We define four key metrics to evaluate CoMOTO and compare our results against several baselines. The human motion predictions are generated in Matlab. All trajectory optimization is performed using TrajOpt~\cite{schulman2014motion}. The experiments are run on a KUKA LBR iiwa R820 Robot in a ROS Gazebo simulation~\cite{hennersperger2017towards}, a visualization of which is shown in Figure \ref{fig:environment}. The coefficients $\alpha_i$ are chosen empirically for optimized performance.

We present three test cases involving close human-robot collaboration, categorized by the behavior of the human:
\begin{flushitemize}
  \item \textbf{Stationary:} Stationary human with robot reaching for an object.
  \item \textbf{Reaching-far:} Human and robot reaching for distant objects.
  \item \textbf{Reaching-near:} Human and robot reaching for closely-positioned objects.
\end{flushitemize}
For each test case, we use $5$ unique human trajectories and a unique nominal robot trajectories. We note that the trajectories within each test case represent different relative initial positions for the human and the robot.
 
\subsection{Human Motion Prediction}
\label{sec:prediction}

We use~\cite{luo2018unsupervised} as the framework for stochastic human motion prediction. The provided code includes ground truth trajectories that are split into training and testing datasets. The trajectories are 3D positions of the human's right arm, $2$ to $3$ seconds long, recorded at 100Hz. The GMM model is trained on 100 trajectories of a human reaching for an object and 100 trajectories of a still human with arm stretched out. Since the dataset only contains recorded trajectories for the right arm (shoulder, elbow, wrist and palm), the remaining human skeleton consisting of the neck, head, torso and left arm is extrapolated using fixed offsets. The same offsets are applied to the mean of the prediction of the right shoulder to generate the predictions of the remaining joints. The covariances for the remaining skeleton are identical to that of the right shoulder.

For each experiment, the ground truth human trajectory is split into two. The first 100 samples (1 second) are used as the observation. A prediction of the remaining human motion is generated based on the observation of those 100 samples. Subsequently, CoMOTO optimizes the robot trajectory such that it is synchronized with the anticipated human motion. We assume that the human remains stationary after reaching the goal.

\subsection{Baselines}
We evaluate CoMOTO, against the following baselines:

\smallskip \noindent
\textbf{Nominal}: the non-adapted nominal trajectory generated by TrajOpt using common costs and constraints, including collision, joint velocity, and joint target constraints. We include the nominal trajectory alone to show how our approach improves this trajectory's performance with respect to the full set of metrics described in Section \ref{sec:metrics}.

\smallskip \noindent
\textbf{Speed-Adjusted}: the nominal trajectory executed with real-time speed adjustment based on human-robot separation distance as described in \cite{lasota2014toward}.

\smallskip \noindent
\textbf{Legible}: the legible motion optimization algorithm of \cite{dragan2013generating}. The baseline implementation uses a legibility cost identical to the one used in our approach. The optimal trajectory is again a linear trajectory in the Cartesian space.

\smallskip \noindent
\textbf{Distant+Visible}: local path optimization using the method presented in \cite{mainprice2011planning}, optimizing for costs based on human-robot separation distance and human visibility. To provide a direct multi-objective optimization comparison to our approach, we use their cost-based optimization to adapt the nominal trajectory, rather than a path generated by a T-RRT planner.

\subsection{Evaluation Metrics}
\label{sec:metrics}

\begin{figure*}[t]
\centering
	\begin{subfigure}[t]{0.95\textwidth}
		\includegraphics[width=\textwidth]{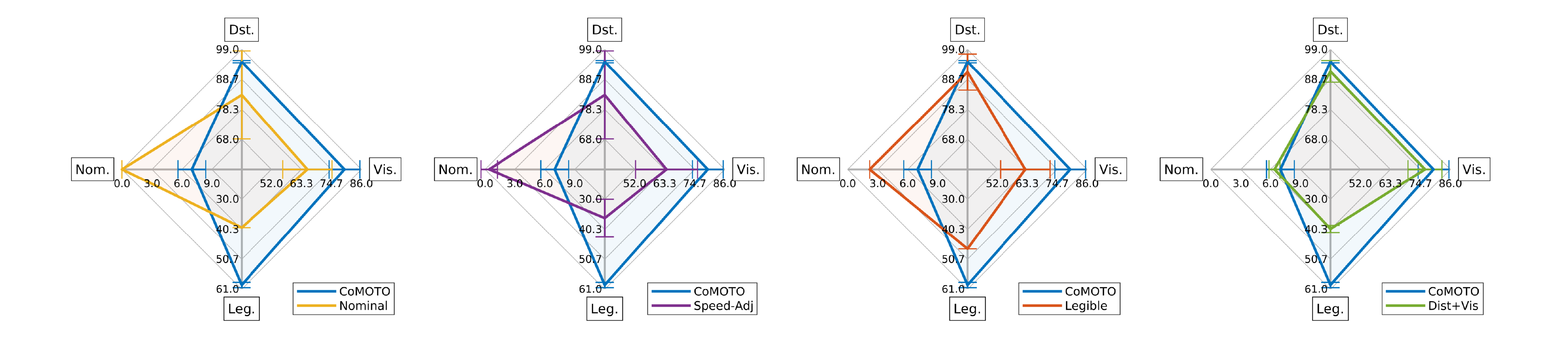}
		\caption{Performance comparison for the stationary scenario.}
		\label{fig:stationary-radar}
	\end{subfigure}
	\begin{subfigure}[t]{0.95\textwidth}
		\includegraphics[width=\textwidth]{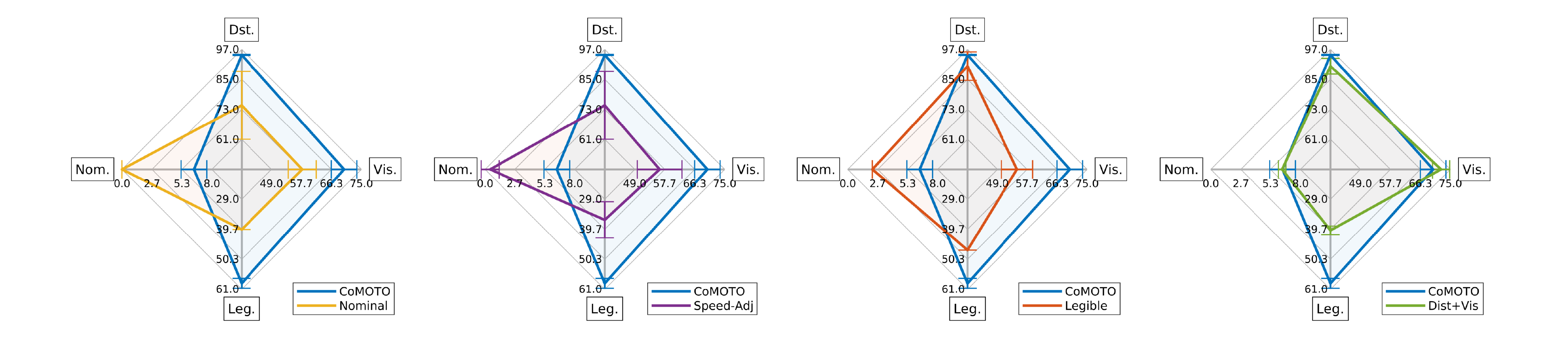}
		\caption{Performance comparison for the reaching-far scenario.}
		\label{fig:reaching-far-radar}
	\end{subfigure}
	\begin{subfigure}[t]{0.95\textwidth}
		\includegraphics[width=\textwidth]{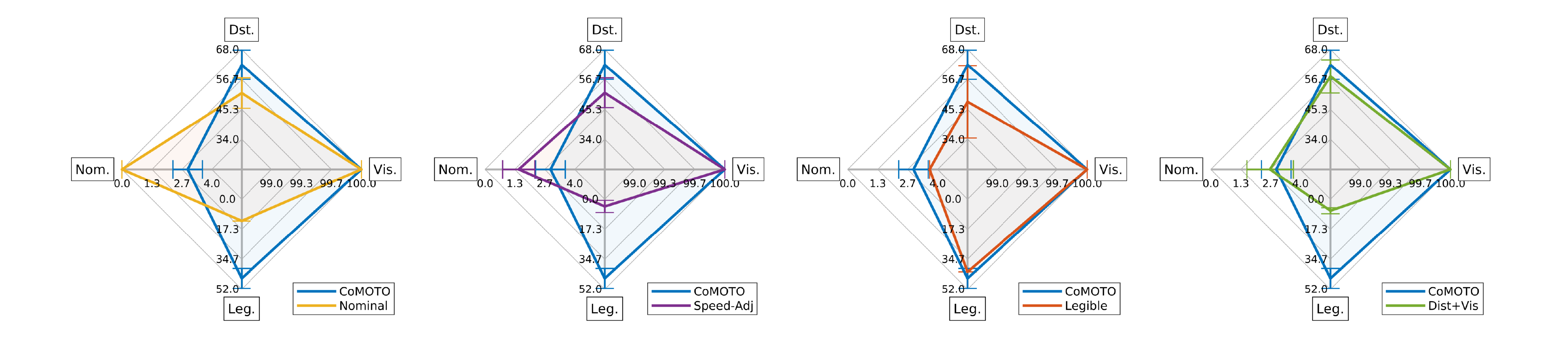}
		\caption{Performance comparison for the reaching-near scenario.}
		\label{fig:reaching-near-radar}
	\end{subfigure}
\caption{Comparison plots between CoMOTO and each baseline with respect to all evaluation metrics (mean $\pm1$ SD)}
\label{fig:radar-all}
\vspace{-0.2cm}
\end{figure*}

We measure the performance of each algorithm according to the following metrics:

\smallskip \noindent
\textbf{Separation distance (Dst.)}: percentage of the trajectory where the separation distance between the robot and the human exceeds 20cm. 

\smallskip \noindent
\textbf{End effector visibility (Vis.)}: percentage of the trajectory where the robot's end effector is within the human's $160^{\circ}$ field of view.  When calculating this metric, we assume the human is looking at their target object.

\smallskip \noindent
\textbf{Legibility (Leg.)}: legibility of the robot's end effector motion to the human observer, calculated as described in~\cite{dragan2013legibility}. 

\smallskip \noindent
\textbf{Deviation from the nominal trajectory (Nom.)}: sum of squared distance between the adapted trajectory and the nominal trajectory.

We note that we utilize the ground truth (as opposed to the predicted) human trajectories to compute the metrics.

\subsection{Results and Discussion}
\begin{table}[t]
\addtolength{\tabcolsep}{-4pt}
\caption{
Each approach's mean $\pm$ 1 SD for all metrics and scenarios,
with the best-performing result shown in bold.  Note that lower values are preferred for the Nominal metric. We denote statistically significant improvements over \textit{all} other methods with * ($p < 0.05$) or ** ($p < 0.01$).}
\vspace{-.2cm}
\label{tab:all-results}
\begin{center}
\renewcommand{\arraystretch}{1.2}
\begin{tabular}{cccccc}
\hline
& Approach & Dst.(\%) & Vis.(\%) & Leg. & Nom. ($m^2$)\\
\hline
\parbox[t]{2mm}{\multirow{5}{*}{\rotatebox[origin=c]{90}{Stationary}}} & CoMOTO & $\mathbf{94.8 \pm 0.4}$ & $\mathbf{79.6 \pm 5.8}$ & $\mathbf{59.5 \pm 0.9}$** & $7.0 \pm 1.4$ \\
& Nominal & $83.3 \pm 15.2$ & $65.6 \pm 9.3$ & $39.7 \pm 0.0$ & \textit{n/a} \\
& Speed-Adj & $83.3 \pm 15.2$ & $64.0 \pm 11.7$ & $36.5 \pm 6.5$ & $\mathbf{0.4 \pm 0.8}$* \\
& Legible & $91.2 \pm 6.2$ & $62.6 \pm 9.4$ & $47.1 \pm 0.0$ & $2.2 \pm 0.0$ \\
& Dist+Vis & $91.4 \pm 3.8$ & $76.4 \pm 6.4$ & $40.2 \pm 1.2$ & $6.4 \pm 0.5$ \\
\hline
\parbox[t]{2mm}{\multirow{5}{*}{\rotatebox[origin=c]{90}{Reaching-far}}} & CoMOTO & $\mathbf{94.8 \pm 0.2}$ & $70.0 \pm 3.7$ & $\mathbf{58.9 \pm 1.8}$** & $6.4 \pm 1.1$ \\
& Nominal & $74.6 \pm 0.1$ & $57.9 \pm 4.0$ & $39.7 \pm 0.0$ & \textit{n/a} \\
& Speed-Adj & $74.6 \pm 13.6$ & $56.2 \pm 6.5$ & $36.3 \pm 6.4$ & $\mathbf{0.5 \pm 0.8}$** \\
& Legible & $90.3 \pm 5.7$ & $54.7 \pm 4.6$ & $47.1 \pm 0.0$ & $2.2 \pm 0.0$ \\
& Dist+Vis & $90.3 \pm 3.2$ & $\mathbf{72.3 \pm 2.5}$ & $40.1 \pm 1.6$ & $6.4 \pm 0.3$ \\
\hline
\parbox[t]{2mm}{\multirow{5}{*}{\rotatebox[origin=c]{90}{Reaching-near}}} & CoMOTO & $\mathbf{62.2 \pm 5.5}$ & $100.0 \pm 0.0$ & $\mathbf{45.7 \pm 5.8}$ & $2.9 \pm 0.7$ \\

& Nominal & $51.5 \pm 5.7$ & $100.0 \pm 0.0$ & $12.5 \pm 0.0$ & \textit{n/a} \\
& Speed-Adj & $51.6 \pm 5.6$ & $100.0 \pm 0.0$ & $4.1 \pm 3.5$ & $\mathbf{1.5 \pm 0.7}$ \\
& Legible & $48.2 \pm 13.7$ & $100.0 \pm 0.0$ & $42.0 \pm 0.0$ & $3.6 \pm 0.0$ \\
& Dist+Vis & $57.8 \pm 6.2$ & $100.0 \pm 0.0$ & $0.0 \pm 1.8$ & $2.6 \pm 1.0$ \\
\end{tabular}
\end{center}
\end{table}

The performance of all algorithms on all metrics can be found in Table \ref{tab:all-results}.  We ran a one-way analysis of variance (ANOVA) with correlated samples for each metric to determine whether differences in our measurements were statistically significant.  Where ANOVA showed a significant difference of trajectory optimization approach on any of our metrics at $p<0.05$, we conducted post tests between the approaches with Tukey's HSD test.  For brevity, we show significant differences in Table \ref{tab:all-results} only for approaches that performed significantly better than \textit{all} other approaches.  We also provide visual comparisons of our approach to each baseline to better show the comparison over all metrics at once in the radar plots of Figure \ref{fig:radar-all}.  We break down the results individually for each test case below.

\subsubsection{Stationary}

We first consider the scenario where the robot must pick an object with a stationary human present in the workspace. The results for this test case are shown in Table \ref{tab:all-results} and Figure \ref{fig:stationary-radar}. Since the motion prediction model was trained on stationary trajectories as well as moving trajectories, the predictions are observed to have minimal motion. With a stationary human, CoMOTO and the Dist+Vis baseline perform comparably on both distance and visibility, with no significant difference. CoMOTO outperforms all other approaches in legibility. 

\subsubsection{Reaching-far}

We next consider the scenario where the human and robot are concurrently reaching for different objects. The results for this test case are shown in Table \ref{tab:all-results} and Figure \ref{fig:reaching-far-radar}. The anticipatory nature of CoMOTO allows it to outperform the Dist+Vis baseline in separation distance and come a close second in visibility. CoMOTO also outperforms all other approaches in legibility. 

\subsubsection{Reaching-near}

Finally, we consider the scenario where the human and robot are concurrently reaching for objects that are close together, providing a situation where high separation distance cannot be maintained. As can be seen in Table \ref{tab:all-results} and Figure \ref{fig:reaching-near-radar}, all methods perform poorly in the distance metric. However, CoMOTO is able to outperform other methods in distance while still maintaining good performance in the other metrics. The Speed-Adjusted baseline is not always able to complete execution of the trajectory since the goal position of the robot may be within its stopping threshold of 6 cm.

\textit{Summary}: Our experiments demonstrate that CoMOTO is a highly adaptable framework that can optimize trajectories to improve factors of safety, comfort, and efficiency across different operating scenarios. CoMOTO significantly improves the performance in separation distance, visibility and legibility over the Nominal baseline. We note that, across all three scenarios, CoMOTO scores consistently better than or equivalent to all baselines both in terms of the distance metric and the visibility metric (see Table \ref{tab:all-results}). This observation suggests that our approach outperforms all the baselines in terms of maintaining a safe and comfortable distance from the human while \textit{simultaneously} making sure that the robot's end effector is visible to the moving human. As one would expect, the Dist+Vis baseline results in slightly better visibility than CoMOTO only in the reaching-far scenario. CoMOTO consistently scores the highest in the legibility metric as well. While CoMOTO doesn't excel in the nominal metric, its performance is comparable to other metrics. Overall, CoMOTO is able to perform well across all metrics while the baselines perform well only in specific metrics or scenarios.

\vspace{-3pt}
\section{Conclusion}
\label{sec:conclusion}

We have presented CoMOTO, a multi-objective trajectory optimization framework for anticipatory human-robot collaboration. 
Our results demonstrate that existing methods tend to specialize and excel on either individual safety, comfort or efficiency metrics, or in particular collaborative scenarios. In comparison, since it simultaneously optimizes for various costs, CoMOTO performs comparably or better across the full set of metrics and scenarios. Future work will explore reactive replanning to tackle unexpected human movements and address computational bottlenecks to enable real-time implementation.


{\small
\bibliographystyle{IEEEtran}
\bibliography{bibliography}
}

\end{document}